\documentclass{article}

\usepackage{PRIMEarxiv}
\usepackage{amsmath}
\usepackage[utf8]{inputenc} 
\usepackage[T1]{fontenc}    
\usepackage{hyperref}       
\usepackage{url}            
\usepackage{booktabs}       
\usepackage{amsfonts}       
\usepackage{nicefrac}       
\usepackage{microtype}      
\usepackage{lipsum}
\usepackage{fancyhdr}       
\usepackage{subcaption} 
\usepackage{graphicx}      
\usepackage[authoryear]{natbib}
\graphicspath{{media/}}     
\usepackage{tabularx,booktabs}
\usepackage{float}
\usepackage{enumitem}

\makeatletter
\renewcommand{\@toptitlebar}{}
\renewcommand{\@bottomtitlebar}{}
\makeatother

\usepackage{setspace}

\title{Languages are Modalities \\
\vspace{6px}
\large Cross-Lingual Alignment via Encoder Injection}

\author{
  Rajan Agarwal$^*$ \\
  University of Waterloo \\
  \texttt{r34agarw@uwaterloo.ca}\\
   \And
  Aarush Gupta$^*$ \\
  Independent \\
  \texttt{hiaarushgupta@gmail.com}
}

\begin{document}
\maketitle
 \def\thefootnote{*}\footnotetext{Denotes equal contribution. Work completed under Cohere Labs Community.}\def\thefootnote{\arabic{footnote}}

\begin{abstract}
Instruction-tuned Large Language Models (LLMs) underperform on low‑resource, non‑Latin scripts due to tokenizer fragmentation and weak cross‑lingual coupling. We present LLINK (\underline{L}atent \underline{L}anguage \underline{I}njection for \underline{N}on-English \underline{K}nowledge), a compute-efficient language-as-modality method that conditions an instruction-tuned decoder without changing the tokenizer or retraining the decoder. First, we align sentence embeddings from a frozen multilingual encoder to the decoder’s latent embedding space at a reserved position via a lightweight contrastive projector. Second, the vector is expanded into $K$ soft slots and trained with minimal adapters so the frozen decoder consumes the signal. LLINK substantially improves bilingual retrieval and achieves 81.3\% preference over the base model and 63.6\% over direct finetuning in LLM-judged Q\&A evaluations. We further find that improvements can be attributed to reduced tokenization inflation and a stronger cross-lingual alignment, despite the model having residual weaknesses in numeric fidelity. Treating low-resource languages as a modality offers a practical path to stronger cross-lingual alignment in lightweight LLMs.
\end{abstract}

\section{Introduction}

Natural languages serve as humanity's primary interface, each encoding unique pragmatics, scripts, and writing systems. A central challenge in machine learning is enabling models to understand, generate, and translate across linguistic variations, to make language models accessible to everyone. However, frontier LLMs today, predominantly trained on English data, demonstrate significant performance degradation on tasks involving low-resource languages, specifically those with non-Latin scripts \citep{petrov2023tokenizer,limisiewicz2023tokenization}.

To mitigate this, current approaches involve in-context learning or moderate finetuning. However, these introduce tokenizer fragmentation, which inflates non-English text into substantially longer sequences, and weak cross-lingual coupling within model representations \citep{petrov2023tokenizer,limisiewicz2023tokenization,ahia2023tokenizationcost,qin2025survey}. Existing solutions to these derivative issues include multilingual pretraining \citep{conneau2020xlmr,xue2021mt5,scao2022bloom} and tokenizer-free byte-level models \citep{xue2022byt5} and character-level encoders \citep{clark2022canine,tay2022charformer}, but carry significant computational and data requirements. Even recent multilingual instruction models still rely on large-scale training and careful tokenizer design \citep{cohere2024ayaexpanse,yang2025qwen25}.
 On the other hand, parameter-efficient finetuning (PEFT) strategies, such as LoRA, IA\textsuperscript{3}, and BitFit \citep{hu2022lora,liu2022ia3,benzaken2022bitfit}, reduce the adaptation cost but still commonly rely on substantial multilingual supervision.

Many languages have sparse web footprints and uneven curation, which makes full multilingual pretraining of LLMs expensive. By contrast, masked-LM encoders trained over many languages handle scarcity relatively well, absorbing monolingual text and share subword/byte structure across scripts, providing strong sequence features. However, these strengths have not translated cleanly to instruction-tuned LLMs. Adding sparse low-resource corpora into continued pretraining or SFT often has minimal effect, causes regressions in English-based evaluation performance, requires more expensive tokenizers, or leads to longer training runs to achieve parity. Therefore, this gap suggests a retrofit path that uses strong external multilingual encoders, rather than trying to make a one-size-fits-all LLM.

\begin{figure}[!htbp]
\centering\includegraphics[width=1\linewidth]{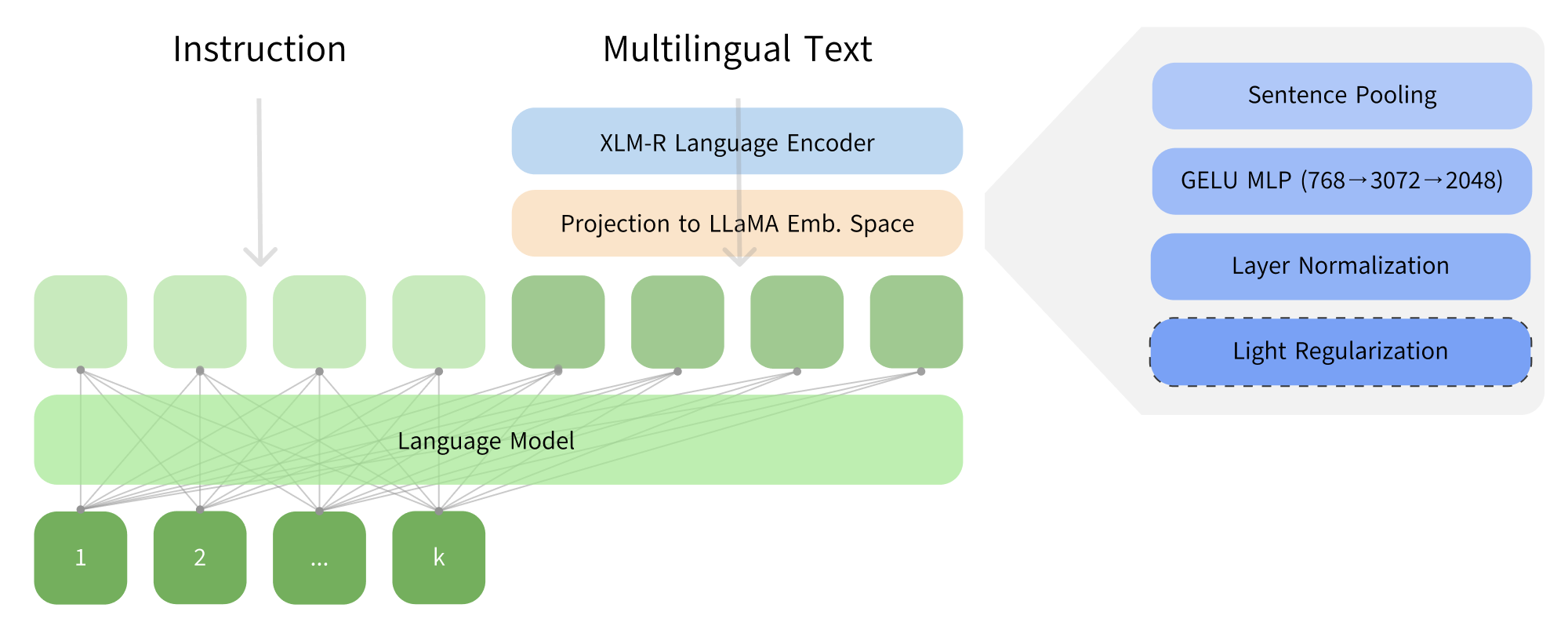}
  \caption{Illustration of LLINK Architecture, passing Multilingual text through a projection model to match LLaMA's embedding space, then to the LLM to produce an output using the translated tokens. Dotted lines represent train-time only.}
\end{figure}
This work treats low-resource source languages as an auxiliary modality for language models. Our paper makes the following contributions:
\begin{itemize}[leftmargin=2em]
    \item \emph{Language‑as‑Modality.} We frame low-resource languages as a modality and inject them into decoder-only LLMs via a compact set of soft slots, bypassing decoder tokenization for non-Latin scripts. This shifts cost from fragmented decoder tokens to a small encoder plus \(K\) slots, yielding up to $\sim3\times$ fewer decoder tokens at prompt time on Khmer while improving cross-lingual quality.
    \item \emph{Contextual teacher alignment.} Between the frozen multilingual encoder and LLM's own hidden state, we apply a contextual teacher alignment at a reserved position, rather than to static token embeddings, providing a stable, context-aware target that strengthens cross-lingual coupling without modifying the tokenizer or decoder weights.
    \item \emph{Usage-enforcing slot objective.} We add a usage-enforcing objective that penalizes the model if replacing injected slots with base embeddings does not worsen loss, making reliance on the foreign signal measurable and trainable. 
\end{itemize}

We train and empirically validate LLINK on Khmer-to-English translation and Q\&A tasks using the ParaCrawl En--Km v2 dataset \citep{banon2020paracrawl,paracrawlSite}. Our method achieves substantial improvements in bilingual retrieval, which we treat as a proxy for evaluating cross-lingual alignment, over direct finetuning baselines. Through LLM-as-Judge pairwise evaluation \citep{zheng2023llmasjudge,liu2023geval}, LLINK-enhanced output are preferred to both the original base model and directly finetuned variants, especially when introduced to tokens out of distribution from the SFT.

\section{Related Work}

\subsection{Tokenizer fragmentation and multilingual inequity.}
Large cross-language disparities arise at the tokenization layer. \citet{petrov2023tokenizer} quantify length inflation up to $15\times$ across languages and show the effect persists for multilingual and byte/character tokenizers. \citet{limisiewicz2023tokenization} analyze vocabulary allocation and overlap, relating them to downstream performance. Tokenizer-free models like ByT5 reduce subword dependence by operating on bytes \citep{xue2022byt5}, and character-level encoders like CANINE and Charformer avoid subword tokenization entirely \citep{clark2022canine,tay2022charformer}, but these approaches induce longer sequences and higher training cost. A complementary line of work adapts vocabularies to reduce cross-lingual inflation without fully retraining the model \citep{yamaguchi2024empirical}. These remedies require retraining with new tokenization or accept efficiency penalties; neither retrofits an existing decoder-only LLM’s tokenizer at inference time.

\subsection{LangBridge \& Multilingual Bridge}
LangBridge \citep{yoon2024langbridge} introduces a lightweight bridge that maps a multilingual encoder’s hidden states (e.g., mT5) into a small sequence of soft-prompt vectors in a frozen decoder-only LM’s input-embedding space. The bridge is trained on English instruction data with a language-modeling objective, so that at inference time non-English inputs are routed through the encoder and injected as continuous prompts, yielding strong zero-shot multilingual reasoning despite English-only supervision. In contrast, LLINK aligns encoder outputs to a reserved decoder hidden state via a two-layer MLP (rather than the input embedding stream) and adds an explicit usage-enforcement objective to ensure the injected slots are consumed during generation.

\subsection{Multimodal Encoder Bridges to LLMs.}
Multimodal stacks such as BLIP-2 \citep{li2023blip2} and LLaVA \citep{liu2023llava} show that small cross-modal encoder injection can provide slot embeddings that the LLM consumes alongside text. BLIP-2 does this with a Q-Former that learns a small set of queries and lets them cross-attend to the encoder features through several transformer layers, adding both parameters and a inference/training cost that grows with the number of queries. LLaVA demonstrates that a simple linear projection trained on instruction-following data suffices to align vision encoder outputs with the LLM's token space, making it much lighter at inference. Speech-to-text and speech-to-LLM systems such as Seamless similarly project non-text modalities into an LLM-consumable representation, reinforcing the view of “modality as just another encoder” \citep{seamless2023}. Prior multilingual bridges typically inject at the embedding stream \citep{yoon2024langbridge}; our method follows this lightweight style but aligns at a reserved decoder hidden state and uses a two-layer MLP plus an explicit usage-enforcement objective, so LLINK keeps a runtime profile closer to LLaVA than to BLIP-2 while improving cross-lingual coupling.

\section{Background}

Modern large language models process text through subword tokenization, typically using Byte-Pair Encoding (BPE) or related algorithms \citep{sennrich2016bpe,kudo2018sentencepiece} trained on predominantly English corpora. This creates severe inefficiencies for non-Latin scripts, with inflation ratios reaching 15$\times$ for some languages \citep{petrov2023tokenizer,lotz2025beyond}, and similar challenges documented for Southeast Asian scripts \citep{ahia2023tokenizationcost}. To quantify this effect for Khmer, we measure tokenization on Khmer-English sentence pairs from ParaCrawl \citep{banon2020paracrawl} using LLaMA-style BPE vocabulary and observe substantial fragmentation.

\begin{figure}[!htbp]
\centering
\begin{subfigure}{\linewidth}
  \centering\includegraphics[width=0.8\linewidth]{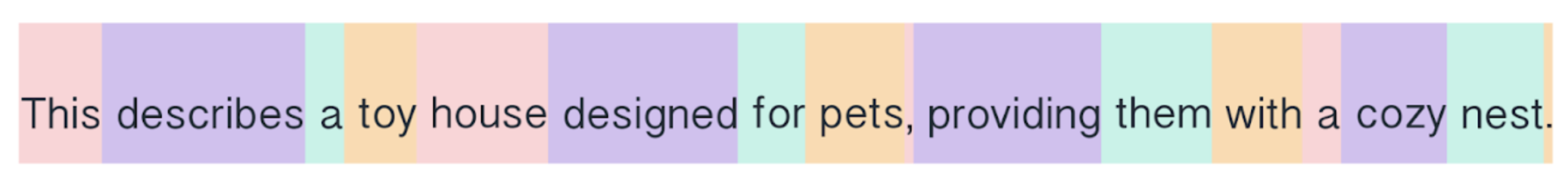}
  \caption{English Sentence, 16 tokens, 0.3 tokens/char}
\end{subfigure}\par\medskip
\begin{subfigure}{\linewidth}
  \centering\includegraphics[width=0.8\linewidth]{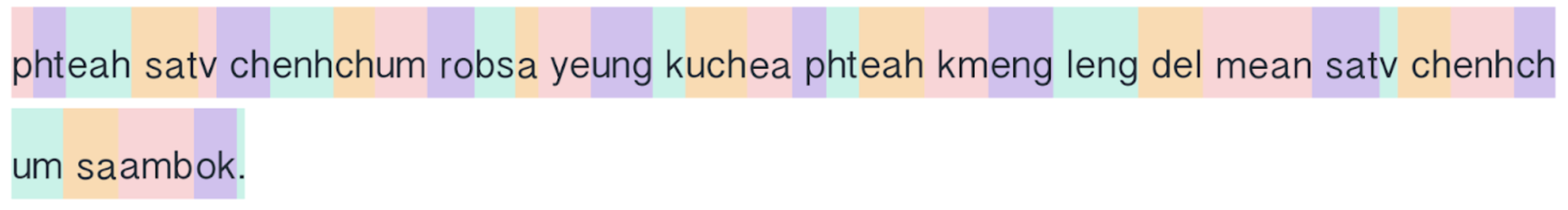}
  \caption{Khmer Latin transliteration Sentence, 35 tokens, 0.5 tokens/char}
\end{subfigure}\par\medskip
\begin{subfigure}{\linewidth}
  \centering\includegraphics[width=0.8\linewidth]{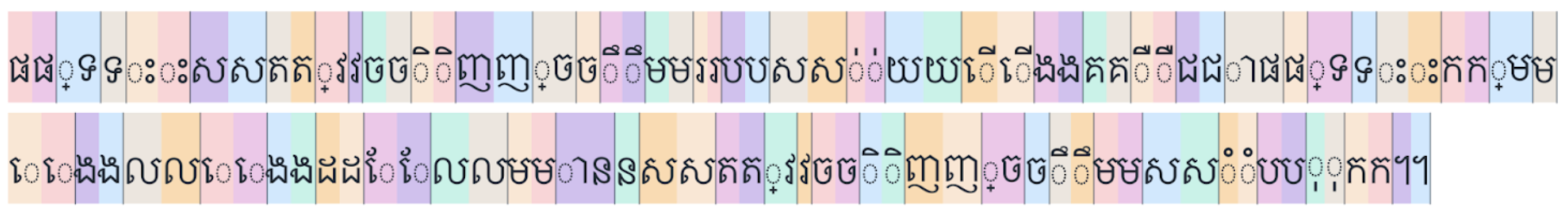}
  \caption{Khmer Sentence, 104 tokens, 1.7 tokens/char}
\end{subfigure}
\caption{Tokenization of the same sentence with the LLaMA-3.2-1B tokenizer — English: 16 tokens (0.3 tok/char); Khmer translit: 35 (0.5); Khmer: 104 (1.7). Dividers on Khmer show duplicate tokens mapping to the same character.}
\label{fig:stacked-sub}
\end{figure}

We examine that processing a Khmer sentence yields approximately $6\times$ more tokens than its English equivalent, and transformer attention compute scales quadratically with sequence length \citep{vaswani2017attention}. A 200-character Khmer sentence might consume 130 tokens, leaving substantially less room for task instructions, few-shot examples, or generated outputs compared to English. The model must learn cross-lingual representations across fragmented tokens, making alignment optimization more difficult.

In Figure~\ref{fig:stacked-sub}, an English sentence tokenizing to 16 tokens expands to 35 tokens when written in Latin transliteration, and further explodes to 104 tokens in native Khmer script using the LLaMA tokenizer. This near-order-of-magnitude difference persists across the distribution. Standard parameter-efficient adaptation methods like LoRA operate on these fragmented token sequences, only inheriting the computational and context-level effects. Even with perfect fine-tuning, the model processes more tokens per forward pass for Khmer inputs compared to English. Our approach sidesteps tokenization at the decoder by treating Khmer as an auxiliary modality, by encoding Khmer text and aligning it to the LLM's latent space through a lightweight projector, \emph{shifting} tokenization overhead to a small, fixed-cost encoder and a few soft slots to reduce decoder-side compute.

\section{Methodology}
We use a two-stage bridge that treats low-resource text as an auxiliary modality. Stage~A learns a small connector that maps a frozen multilingual encoder’s sentence representation into the LLM’s latent space at a reserved position—no tokenizer changes, no heavy retraining. Stage~B then exposes this signal to the decoder via a few soft slots and lightly tunes small modules so the model actually relies on it during generation. This approach circumvents tokenization inflation while preserving the base model's weights. We provide the architectural and training details in the following sections; related connector-style approaches are discussed in Section~5.

\subsection{Stage A: Contrastive Alignment}
We first build a single, deterministic “foreign representation’’ at a reserved slot the decoder can read. A frozen XLM-R encodes a Khmer sentence and we mask–mean pool the token states to a sentence vector $\mathbf{z}_F\!\in\!\mathbb{R}^{768}$. On the English side, we append a reserved token at the end of the user instruction and take the final hidden state at that position as the teacher target ($h_E$). We use the following prompt template (Stage A):
 \texttt{User:<instruction><foreign\_emb> Assistant:...} .
 This fixes ($h_E$) to a known context position and makes the target prompt-dependent but decoder-stable. As a result, final latent state at that position is the teacher target $\mathbf{h}_E\!\in\!\mathbb{R}^{2048}$. The input-embedding row for \texttt{<foreign\_emb>} is zero-initialized; under RMSNorm (no bias) \citep{zhang2019rmsnorm} a zero vector keeps the slot neutral before alignment as residual self-attention moves the state only via context.

A lightweight projector MLP $g$ maps encoder representations into the decoder space:
$g:\;768\!\rightarrow\!3072\!\rightarrow\!2048$ with Linear+GELU \citep{hendrycks2016gelu}, dropout 0.10, and a final LayerNorm \citep{ba2016layernorm}. We set $\mathbf{p}_F=g(\mathbf{z}_F)$ and update \emph{only} $g$, ensuring the multilingual encoder and decoder stay frozen. The training objective is
\[
\mathcal{L} \;=\; \tfrac{1}{2}\big[\mathrm{NCE}(\mathbf{p}\!\rightarrow\!\mathbf{h})+\mathrm{NCE}(\mathbf{h}\!\rightarrow\!\mathbf{p})\big]\;+\; \lambda_{\text{dir}}\mathcal{L}_{\text{dir}} \;+\; \lambda_{\text{norm}}\mathcal{L}_{\text{norm}}.
\]
For the symmetric InfoNCE \citep{oord2018cpc}, for each parallel pair we treat $(\mathbf{p}_F,\mathbf{h}_E)$ as the positive, use in-batch negatives, and augment the denominator with hard negatives. \citep{he2020moco,radford2021clip}. For a strong selection of negatives, we maintain a 32768-item fp16 FIFO queue of cosine-normalized teacher vectors and, per step, select the 256 hardest by cosine similarity.

We add two light regularizers. First, a direction term $\mathcal{L}_{\text{dir}}=\|{\mathbf{p}}_F-{\mathbf{h}}_E\|^2$ with L2 regularization and vector normalization, as well as weight $0.05$, acting as an anchor to push alignment beyond what InfoNCE alone enforces. Second, a log-norm matching term $\mathcal{L}_{\text{norm}}=\big(\log\|\mathbf{p}_F\|-\log\|\mathbf{h}_E\|\big)^2$ with weight $0.02$, which keeps magnitudes comparable and prevents exploding or vanishing norms. The result is a compact vector at a known position that reliably summarizes the foreign sentence in the decoder’s own space.

\subsection{Stage B: Multi-Token Injection with Usage Enforcement}
However, the decoder may still disregard the aligned foreign representation without explicit training signals. Stage~B aims to encourage usage of the vector by expanding it into $K$ soft tokens (akin to soft prompt tuning \citep{lester2021prompttuning}) and adding training signals so the model learns to rely on them during generation.

In LLINK, we expand the Stage~A vector into \(K\) slot embeddings and replace the single \texttt{\textless foreign\_emb\textgreater} with reserved tokens \texttt{\textless f0\textgreater}–\texttt{\textless fk-1\textgreater}. These slots live in the context like ordinary tokens and can be attended at every layer, giving enough capacity to carry longer or denser content than a single 2048-d point. To expand the vector, we unit-normalize, apply a non-affine LayerNorm, and multiply by a learned scalar initialized to the median norm of the base embedding matrix; the $K=8$ slots are inserted after the instruction as \texttt{<f0>…<f7>}. We inject by computing base embeddings and overwriting the rows at the reserved token positions so the decoder sees them directly.

Then, to teach the model to read the slots, we apply low-rank LoRA to attention and MLP projections and train it jointly with the scale, adapter, and expander. This keeps the Stage~A projector and all other base weights remain frozen. We synthetically generate Q\&A prompts that use a \texttt{\textless foreign\_emb\textgreater}  vector and reply in English. Four task templates were included in this distribution, including \texttt{bullet\_pointify}, \texttt{translate\_to\_english}, \texttt{summarize\_in\_english} and \texttt{qa\_about\_text}. 

Prompts are tokenized with \texttt{<f0>}–\texttt{<f7>} and the pipeline mirrors inference, first encoding Khmer with XLM-R, projecting its embeddings with scale/adapter, expand to \(K\) slots and then decode. Providing features alone does not guarantee usage, so we add a lightweight usage-contrast. Every third step we compute the supervised fine-tuning loss with injected slots, \(\mathcal{L}_{\text{SFT}}\), and a “zeroed’’ loss \(\mathcal{L}_{\text{zero}}\) where the reserved positions are restored to the original embeddings. We penalize cases where removal helps:
\[
\mathcal{L}_{\text{contrast}} \;=\; 0.05\,\max\!\bigl(0,\ \mathcal{L}_{\text{SFT}} - \mathcal{L}_{\text{zero}}\bigr).
\]
Two small alignment auxiliaries anchor the slots to the Stage~A target space without dominating training. A unit-vector matching term increases the cosine similarity to the teacher slot, and an InfoNCE loss with weight 0.01 against the same teacher to prevent drift. The total objective is \(\mathcal{L}_{\text{SFT}} + \mathcal{L}_{\text{contrast}} + \) (auxiliaries). In practice this shifts the model from paraphrasing around the slots to actually using them, while keeping changes to the base LLM minimal.

\section{Experimental Setup}

\subsection{Data}

We use ParaCrawl v2 English–Khmer \citep{banon2020paracrawl} for all experiments. We truncate Khmer strings to at most 256 characters, and take a 140k English-Khmer pair subset from the dataset, dividing it into 100k training pairs for Stage~A and 40k holdout for retrieval evaluation.

Stage~B requires instruction-following examples that use the injection pipeline. We synthesize instruction-following examples from parallel pairs using a LLaMA‑3 70B instruction‑tuned model \citep{LLaMA3herd2024}. For each Khmer sentence, the model is conditioned on the reference English translation so that targets are anchored to ground truth rather than model output. After filtering for non-empty inputs/targets, presence of the reserved token \texttt{\textless foreign\_emb\textgreater} in the prompt, and Khmer length between 12-256 characters, we select the Stage~B set with 40k training and 2k validation examples.

\subsection{Baselines and Evaluation}

The base model is LLaMA-3.2-1B-Instruct with no adaptation, processing Khmer directly through its BPE tokenizer and suffering from the measured 6.5$\times$ fragmentation. Direct fine-tuning applies LoRA (rank 16, alpha 16, same configuration as LLINK's LoRA component) to the base model on instruction data where Khmer is tokenized normally, representing standard parameter-efficient adaptation.

For evaluation, we first use bilingual retrieval to represent cross-lingual alignment quality, following recent CLIR-style evaluations for multilingual LLMs \citep{goworek2025bridging}.
 Given $N$ English-Khmer pairs, we encode all Khmer sentences through our pipeline and all English sentences through the teacher position (append \texttt{<foreign\_emb>}, extract hidden state). We compute cosine similarity and report Recall at ranks 1, 5, and 10, plus Mean Reciprocal Rank (MRR) and Mean Rank. This tests whether aligned representations enable correct matching, which correlates with translation quality. To test generations, we use a LLaMA‑3 70B instruction model as a judge \citep{LLaMA3herd2024}, following LLM‑as‑Judge methodology \citep{zheng2023llmasjudge,liu2023geval}, with anonymized pairwise comparisons. For each of 500 test examples, we generate outputs from two systems and record full win/loss/tie breakdowns.
\section{Results}

\subsection{Retrieval alignment}
We evaluate Khmer to English alignment on a held-out set of 1{,}024 parallel pairs. For each Khmer sentence, we compare its normalized LLINK projection to the normalized teacher vectors extracted at \texttt{\textless foreign\_emb\textgreater} for all English sentences and rank the gold target among 1{,}024 candidates. We report Recall@k (R@k), mean reciprocal rank (MRR), and mean rank. Table~\ref{tab:retrieval} shows large gains over a direct fine-tune, with R@1 improving from 0.104 to 0.450 ($\sim$4.3$\times$), and a sharp drop in mean rank. Stage A provides the primary improvement by bypassing tokenization inflation and directly aligning to the decoder's representation space, reducing false matches.

We measure end-to-end quality with anonymized A/B comparisons and an LLM judge (LLaMA~3.1~70B Instruct). For each prompt, the judge sees two model outputs in random order and the human reference translation (for verification), then returns \{win, loss, tie\}. Preference is wins/(wins+losses). We bucket prompts into two task types: (i) Q\&A about the foreign content, and (ii) content understanding (translation, summary, paraphrase, title).

\begin{table}[t]
\centering
\begin{tabular}{lccccc}
\toprule
Method & R@1 & R@5 & R@10 & MRR & Mean Rank \\
\midrule
Direct fine-tune & 0.104 & 0.248 & 0.352 & 0.160 & 24.7 \\
LLINK (Stage~A)  & 0.430 & 0.706 & 0.819 & 0.642 & 3.8 \\
LLINK (Full)     & \textbf{0.450} & \textbf{0.724} & \textbf{0.835} & \textbf{0.660} & \textbf{3.4} \\
\bottomrule
\end{tabular}
\vspace{6pt}
\caption{Bilingual retrieval (R@k, MRR, mean rank) on n = 1,024 held-out Khmer–English pairs.}
\label{tab:retrieval}
\end{table}

\begin{table}[t]
\centering
\begin{tabular}{lccccc}
\toprule
Bucket & Comparison & Wins \% & Losses \% & Ties \% & Preference \\
\midrule
Content Understanding (n=500) & LLINK vs.\ Base & 69 & 11 & 20 & \textbf{86.3\%} \\
                & LLINK vs.\ Fine-tune & 45 & 23 & 32 & \textbf{66.2\%} \\
\midrule
Q\&A (n=500)    & LLINK vs.\ Base  & 48 & 15 & 36 & \textbf{76.2\%} \\
    & LLINK vs.\ Fine-tune & 39 & 25 & 36 & \textbf{60.9\%} \\
\bottomrule
\end{tabular}
\vspace{6pt}
\caption{LLM-as-judge evaluation with selected permutations (judge sees the human reference; preference excludes ties). Judge: LLaMA~3.1~70B Instruct.}
\label{tab:judge}
\end{table}

\begin{table*}[!t]
\centering
\begin{tabularx}{\textwidth}{@{}>{\raggedright\arraybackslash}X >{\raggedright\arraybackslash}X@{}}
\toprule
\textbf{Positive cases (LLINK preferred)} & \textbf{Negative cases (baseline preferred)} \\
\midrule

{\small
\textbf{Q\&A}\par
\emph{Question:} What policy is stated?\par
\emph{Translated Khmer:} We do not share your contact information unless you consent.\par
\emph{Base:} (untranslatable Khmer; unhelpful) \par
\emph{LLINK:} The company will not disclose personal data without consent.
}
&
{\small
\textbf{Content understanding—translation}\par
\emph{Task:} Translate to English.\par
\emph{Translated Khmer:} We have four perfect crafts to play with.\par
\emph{Base:} We have four good instruments to play with.\par
\emph{LLINK:} This suggests a variety of games available for players.
}
\\[0.75em]

{\small
\textbf{Q\&A}\par
\emph{Question:} When will classes resume?\par
\emph{Translated Khmer:} Classes resume Monday.\par
\emph{Base:} School will reopen soon.\par
\emph{LLINK:} Classes will resume on Monday.
}
&
{\small
\textbf{Q\&A}\par
\emph{Question:} What does the text say about capacity?\par
\emph{Translated Khmer:} Capacity: 30\,MW.\par
\emph{Base:} Capacity: 30\,MW.\par
\emph{LLINK:} This indicates the power rating of the device, which is 1.5\,kW.
}
\\[0.75em]

{\small
\textbf{Content understanding—translation}\par
\emph{Task:} Translate to English.\par
\emph{Translated Khmer:} There is a specific door type.\par
\emph{Base:} This indicates a specific model number.\par
\emph{LLINK:} This indicates a specific type of door.
}
&
{\small
\textbf{Content understanding—summary}\par
\emph{Task:} Summarize in English.\par
\emph{Translated Khmer:} Submit the application by Friday.\par
\emph{Base:} Submit the application.\par
\emph{LLINK:} Send the form this weekend.
}
\\
\bottomrule
\end{tabularx}
\caption{Side-by-side qualitative cases used in the LLM-as-judge evaluation with judge LLaMA~3.1~70B Instruct; comparisons are blinded A/B with access to the human reference for verification. LLINK vs Base, LLINK averages to a 81.3\% preference. LLINK vs Fine-Tune, 63.6\% preference.}
\label{tab:qualitative_sxs}
\end{table*}

Gains are largest on Q\&A, where the slots act like a grounded summary the decoder can copy facts from. On content-understanding tasks, LLINK improves precision (fewer mixed-script or off-topic outputs) but will paraphrase rather than translate literally. This behavior is expected, as the encoding process, project it to another space and use those vectors, which would preserve meaning but not specific words or numbers. Stage~B’s usage-contrast helps, but lexical exactness can still lag when the underlying reference contains uncommon terms. These show in many forms, a few notable examples being unit slips (kW vs MW), category substitution (``games'' vs ``instruments''), and occasional over-summarization.

\section{Analysis}

\subsection{Understanding LLINK's Effectiveness}

The dramatic performance gap between Stage A alone (R@1: 0.104 to 0.430) and the full model (to 0.450) reveals that tokenization fragmentation is the dominant bottleneck for cross-lingual understanding. By replacing 104 fragmented Khmer tokens with 8 semantic slots, we inherently change how the decoder processes foreign text. Now, instead of attending over incomprehensible fragments, it sees coherent semantic units. Unlike static embedding mapping approaches, we align to hidden states at a reserved position \textit{after} the decoder has processed the English context. This provides a richer, context-aware target that already encodes task-relevant information and expected answer formats. This method teaches the projection to produce representations that fit naturally into the decoder's existing computational flow. The frozen decoder acts as an implicit regularizer, preventing the aligned representations from drifting into decoder-incompatible regions of the hidden space.

This architectural choice creates a natural trade-off. Consider LLINK as a lossy semantic compression, such that variable-length sequences become fixed K-dimensional representations. This explains both our strong performance on meaning-preservation tasks (Q\&A, retrieval) and systematic failures on surface-form tasks (numeric precision, exact translation). The observed confusion between "30 MW" and "1.5 kW" reflects how multilingual encoders represent numbers on logarithmic scales where these values are semantically proximate.

The usage-enforcement objective also reveals  that even well-aligned representations can be ignored without explicit training pressure. The decoder's strong English priors resist foreign signals, preferring to paraphrase around unknown content rather than utilize it directly. This resistance might explain why previous multilingual bridging attempts showed limited success without extensive adaptation.

\subsection{Computational Trade-offs}

LLINK achieves approximately 3× reduction in decoder tokens in our experiments, by shifting computational burden from the decoder to a one-time encoding cost. This trade-off favors scenarios where encoder cost amortizes across multiple uses (batch processing, caching, or repeated queries) but may not benefit single-pass translation.

The preference gaps in judge evaluation (81.3\% vs base, 63.6\% vs fine-tune) suggest the base model produces mixed-script nonsense, fine-tuning learns brittle pattern matching on fragments, while LLINK maintains semantic coherence but loses lexical precision. This creates a taxonomy, where semantic understanding tasks benefit from LLINK's approach, while applications requiring exact reproduction may need augmentation with copying mechanisms or hybrid strategies.

\subsection{Future Work}

While LLINK demonstrates promise for Khmer-English tasks, several directions merit exploration:

\textbf{Scalability across languages and models.} Testing on typologically diverse languages (Arabic RTL, Chinese logographic, Swahili agglutinative) would validate generalization. Similarly, scaling to larger decoders (7B, 13B) requires investigation. It may be hypothesized that stronger English priors in larger models will necessitate adjusted usage enforcement or increased K.

\textbf{Dynamic slot allocation.} Our fixed K=8 represents a compromise across tasks. Adaptive allocation based on input length, complexity, or entropy could improve efficiency. For example, simple queries might need only K=2-4, while technical documents benefit from K=12-16. A lightweight classifier could predict optimal K at inference time.

\textbf{Hybrid precision mechanisms.} To address numeric and entity errors, we envision augmenting LLINK with specialized pathways: (1) a copying mechanism that preserves exact strings when detected, (2) dedicated slots for numbers that bypass semantic compression, or (3) attention supervision that encourages direct slot-to-output correspondence for critical tokens.

\textbf{Many-to-many language bridging.} Current work assumes English as the target. Extending to arbitrary language pairs requires either training pairwise projectors or learning a universal interlingual space. The latter is appealing but may sacrifice language-specific nuances.

\section{Conclusion}
LLINK frames low-resource languages as a modality for decoder-only LLMs, aligning compact foreign representations to a place the decoder already understands and then ensuring that this signal is actually used. This design circumvents tokenization inflation and delivers robust semantic coupling with modest engineering and compute. The same design also explains how compressed, slot-based injection favors meaning over surface form and can lose exact numerals and lexical detail. The analysis above identifies why this happens, when it matters, and how to mitigate it. 

Treating low-resource, non-Latin scripts as a modality offers a compute-efficient path to improved cross-lingual behavior without retraining tokenizers or decoders, potentially broadening access for underserved languages. At the same time, mis-translations that alter numbers, units, or named entities can have outsized impact. With copy-aware training, mild structural capacity in the slots, diversified teacher targets, and numeracy-focused supervision, we anticipate maintaining LLINK’s efficiency while closing the gap on lexical fidelity; we intend that this work helps move closer to practical, small-model cross-lingual systems that serve languages underrepresented in current tokenizers and pre-training.

\section*{Acknowledgments}
We thank Cohere Labs for their support, the Modal team for compute, the ParaCrawl project for parallel data, and the XLM-R and LLaMA teams for open-source models that made this work possible.

\newpage
\bibliography{references}

\newpage

\newcommand{\GenRow}[2]{\textbf{#1} & \parbox[t]{0.82\textwidth}{\raggedright #2}\\}

\newenvironment{GenCase}[2]{%
  \begin{table*}[t]
  \small
  \setlength{\tabcolsep}{6pt}
  \renewcommand{\arraystretch}{1.15}
  \begin{tabularx}{\textwidth}{@{}lX@{}}
  \toprule
  \textbf{ID} & #1\\
  \textbf{Task} & #2\\
}{%
  \bottomrule
  \end{tabularx}
  \end{table*}
}

\newcommand{\sys}[1]{\textsc{#1}}

\appendix

We release the training and inference code at \hyperlink{github.com/rajansagarwal/llink}{https://github.com/rajansagarwal/llink}.

\section{Tokenization Analysis}

\begin{figure}[!htbp]
\centering\includegraphics[width=1\linewidth]{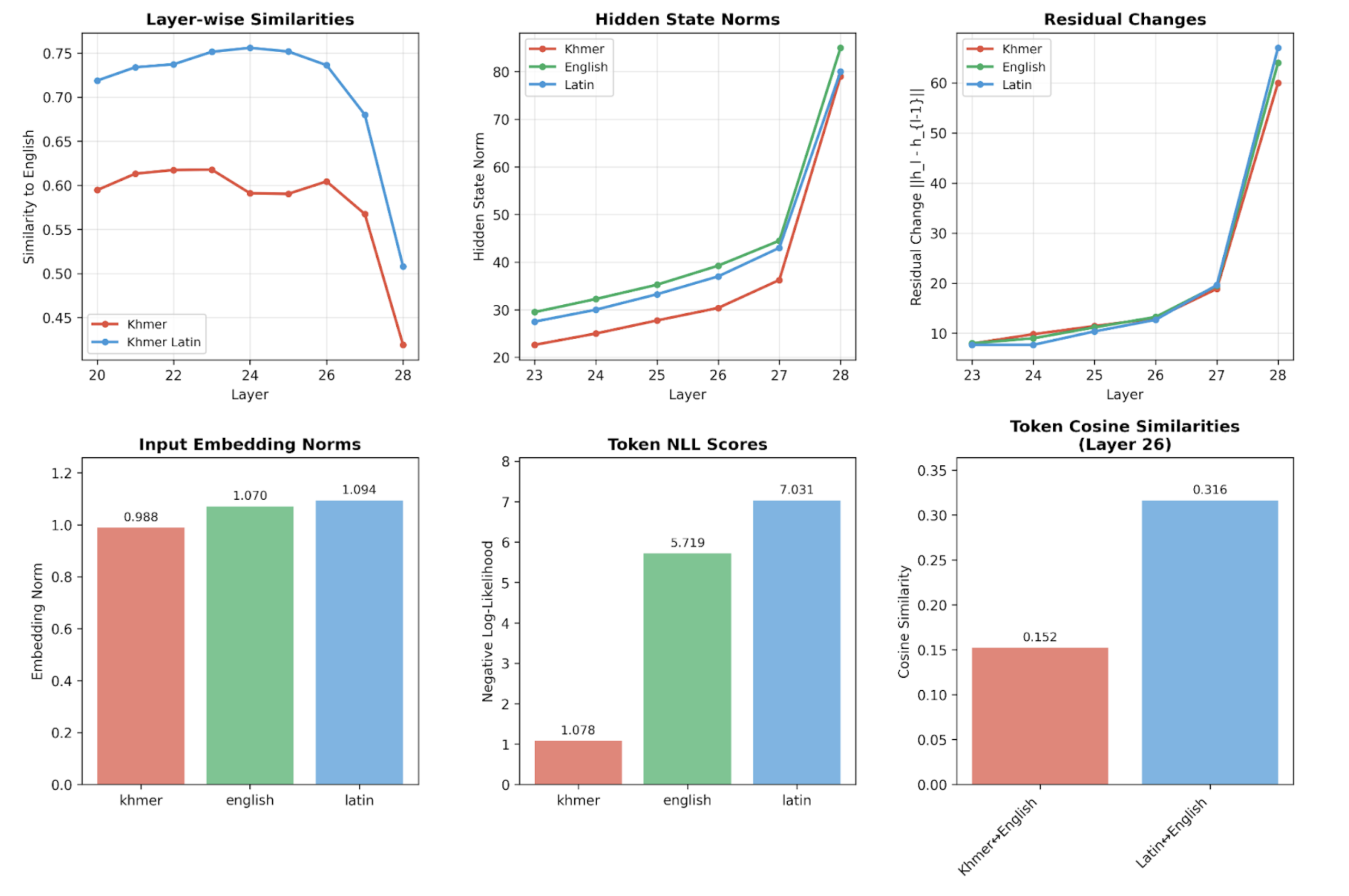}
  \caption{Analysis of fine-tuned representations with Khmer LLaMA 3.2 tokenization. The top three charts present layer-wise similarities, hidden state norms and residual changes. The bottom three charts present input embedding norms, token NLL scores and cosine similarities between Khmer, Khmer Latin transliteration and English translations of the same text.}
\end{figure}

LLaMA 3.2 shows disparate token-level NLL scores across languages: Khmer text exhibits a mean NLL of 1.078, while English and Latin transliterations achieve 5.719 and 7.031, respectively. This 5.3-fold advantage for Khmer represents an asymmetry in predictive difficulty. The substantially lower NLL for Khmer-tokenized text suggests that the model's tokenizer, despite being primarily trained on Latin-script corpora, has developed efficient compression strategies for the Khmer script, which we note in sparser token sequences that align more naturally with the model's learned distributional patterns.

The disparity in cosine similarity of the different scripts indicates that while Khmer tokenization produces sequences that are easier to predict, these sequences occupy more distant regions of the representation space relative to semantically equivalent English text. The Latin transliteration, despite a higher NLL, produces representations that share a greater geometric similarity with English, likely due to shared character-level and phonetic features.

Through the middle layers (20--26), we see progressive semantic alignment, where Khmer maintains cosine similarities with English between 0.59 and 0.62, while Latin achieves 0.72 to 0.76. The lower norm for Khmer embeddings suggests that Khmer tokens, on average, occupy less extreme positions in the initial embedding space, potentially reflecting the tokenizer's treatment of the script as a more compact symbolic system.

\section{More LLINK Generations}
\label{app:generations}

\begin{table}[H]
\small
\setlength{\tabcolsep}{6pt}
\renewcommand{\arraystretch}{1.15}
\begin{tabularx}{\linewidth}{@{}l>{\raggedright\arraybackslash}X@{}}
\toprule
\textbf{Task} & Content understanding — translate \\
\textbf{Reference (En)} & Good intentions for 2021? \\
\textbf{Base} & 2021 (untranslatable Khmer) \\
\textbf{Fine-tune} & The 2021 Myanmar military coup. \\
\textbf{LLINK} & \textbf{This suggests a positive outlook for the upcoming year.} \\
\bottomrule
\end{tabularx}
\end{table}

\begin{table}[H]
\small
\setlength{\tabcolsep}{6pt}
\renewcommand{\arraystretch}{1.15}
\begin{tabularx}{\linewidth}{@{}l>{\raggedright\arraybackslash}X@{}}
\toprule
\textbf{Task} & Content understanding — summarize \\
\textbf{Reference (En)} & The store sells books, magazines, and newspapers. \\
\textbf{Base} & Store sells items. \\
\textbf{Fine-tune} & The store has reading materials. \\
\textbf{LLINK} & \textbf{The store offers books and publications.} \\
\bottomrule
\end{tabularx}
\end{table}

\begin{table}[H]
\small
\setlength{\tabcolsep}{6pt}
\renewcommand{\arraystretch}{1.15}
\begin{tabularx}{\linewidth}{@{}l>{\raggedright\arraybackslash}X@{}}
\toprule
\textbf{Task} & Q\&A about foreign text (numeric weakness) \\
\textbf{Reference (En)} & The distance is 25 kilometers. \\
\textbf{Prompt (En)} & How far is it? \\
\textbf{Base} & (untranslatable Khmer; no answer) \\
\textbf{Fine-tune} & It is far away. \\
\textbf{LLINK} & \textbf{The distance is approximately 10 miles.} \\
\bottomrule
\end{tabularx}
\end{table}

\begin{table}[H]
\small
\setlength{\tabcolsep}{6pt}
\renewcommand{\arraystretch}{1.15}
\begin{tabularx}{\linewidth}{@{}l>{\raggedright\arraybackslash}X@{}}
\toprule
\textbf{Task} & Content understanding — translate \\
\textbf{Reference (En)} & Parking is available on the second floor. \\
\textbf{Base} & (untranslatable Khmer; no answer) \\
\textbf{Fine-tune} & There is parking in the building. \\
\textbf{LLINK} & \textbf{Cars go on the second floor.} \\
\bottomrule
\end{tabularx}
\end{table}

\begin{table}[H]
\small
\setlength{\tabcolsep}{6pt}
\renewcommand{\arraystretch}{1.15}
\begin{tabularx}{\linewidth}{@{}l>{\raggedright\arraybackslash}X@{}}
\toprule
\textbf{Task} & Content understanding — bullet pointify \\
\textbf{Reference (En)} & Requirements: valid ID, proof of address, application form. \\
\textbf{Base} & Requirements are (untranslatable Khmer) \\
\textbf{Fine-tune} & • ID • Successful • Form \\
\textbf{LLINK} & \textbf{• Proper identification • Name • City} \\
\bottomrule
\end{tabularx}
\end{table}

\begin{table}[H]
\small
\setlength{\tabcolsep}{6pt}
\renewcommand{\arraystretch}{1.15}
\begin{tabularx}{\linewidth}{@{}l>{\raggedright\arraybackslash}X@{}}
\toprule
\textbf{Task} & Content understanding — summarize \\
\textbf{Reference (En)} & The course includes lectures, group work, and a final exam. \\
\textbf{Base} & (untranslatable Khmer) \\
\textbf{Fine-tune} & The course will test you on content. \\
\textbf{LLINK} & \textbf{Students attend lectures and do projects and tests.} \\
\bottomrule
\end{tabularx}
\end{table}

\begin{table}[H]
\small
\setlength{\tabcolsep}{6pt}
\renewcommand{\arraystretch}{1.15}
\begin{tabularx}{\linewidth}{@{}l>{\raggedright\arraybackslash}X@{}}
\toprule
\textbf{Task} & Q\&A about foreign text (numeric error) \\
\textbf{Reference (En)} & The building was constructed in 1985. \\
\textbf{Prompt (En)} & When was the building built? \\
\textbf{Base} & 1985 \\
\textbf{Fine-tune} & 1985 \\
\textbf{LLINK} & \textbf{The building was built in 1980s} \\
\bottomrule
\end{tabularx}
\end{table}

\begin{table}[H]
\small
\setlength{\tabcolsep}{6pt}
\renewcommand{\arraystretch}{1.15}
\begin{tabularx}{\linewidth}{@{}l>{\raggedright\arraybackslash}X@{}}
\toprule
\textbf{Task} & Q\&A about foreign text \\
\textbf{Reference (En)} & The fee is \$50 per person. \\
\textbf{Prompt (En)} & How much does it cost? \\
\textbf{Base} & (untranslatable Khmer; no answer) \\
\textbf{Fine-tune} & There is a fee of \$50. \\
\textbf{LLINK} & \textbf{The cost is \$5 each.} \\
\bottomrule
\end{tabularx}
\end{table}

\end{document}